\documentclass[report, 11pt]{article}

\usepackage{graphicx}
\usepackage{float}
\usepackage{tabularx}
\usepackage[margin=1.1in]{geometry}
\usepackage[numbers]{natbib}
\usepackage{adjustbox}
\usepackage{hyperref}
\usepackage{setspace}

\newcommand\TableCite[1]{%
  \citeauthor{#1}~\citeyear{#1}~[\citenum{#1}]}
  
  \providecommand{\keywords}[1]
  {
  	\small	
  	\textbf{\textit{Keywords---}} #1
  }

\begin{document}
\sloppy

\title{Reality-assisted evolution of soft robots through large-scale physical experimentation: a review} 

\author{Toby~Howison\thanks{th533@cam.ac.uk}, Simon~Hauser, Josie~Hughes and Fumiya~Iida \thanks{Bio-Inspired Robotics Lab, University of Cambridge, UK.}}

\maketitle
\doublespacing

\begin{abstract}
In this review we introduce the framework of reality-assisted evolution to summarize a growing trend towards combining model-based and model-free approaches to improve the design of physically embodied soft robots. \emph{In silico}, data-driven models build, adapt and improve representations of the target system using real-world experimental data. By simulating huge numbers of virtual robots using these data-driven models, optimization algorithms can illuminate multiple design candidates for transference to the real world. \emph{In reality}, large-scale physical experimentation facilitates the fabrication, testing and analysis of multiple candidate designs. Automated assembly and reconfigurable modular systems enable significantly higher numbers of real-world design evaluations than previously possible. Large volumes of ground-truth data gathered via physical experimentation can be returned to the virtual environment to improve data-driven models and guide optimization. Grounding the design process in physical experimentation ensures the complexity of virtual robot designs does not outpace the model limitations or available fabrication technologies. We outline key developments in the design of physically embodied soft robots under the framework of reality-assisted evolution.
\end{abstract}


\section{Introduction}
From simple cellular organisms to the most complex lifeforms, biological systems exhibit extraordinary levels of diversity over a large spatio-temporal scale. The apparently coherent behavioral repertoires observed in nature are increasingly explained using an \emph{embodied} view of intelligence that focuses on the closely coupled interactions between the brain, body and environment \cite{pfeifer2007self, clark2008supersizing}. Understanding and harnessing these embodied interactions, therefore, is a key step in building truly lifelike artificial systems. 

In this context the field of \emph{soft} robotics is particularly interesting. Soft robots can generally be defined as having highly deformable bodies or elements, often constructed using unconventional (in the robotics sense) materials with highly non-linear properties \cite{pfeifer2012challenges,nurzaman2014soft,rus2015design,laschi2016soft}. They can be further characterized by their uniquely complex and sustained environmental interactions, for example in terrestrial or aquatic environments \cite{corucci2018evolving}. Given this, the methodologies on which conventional robotics science was developed---e.g. rigid-body kinematics and dynamics---are often inapplicable when applied to soft robots. Instead, significant research has focused on how to utilize the characteristics of soft robots in an embodied framework. By harnessing the mechanical intelligence of soft bodies, researchers hope to develop robotic systems with comparable performance to biological systems \cite{kim2013soft}. 

While the promise of soft embodied intelligence and soft robotics is well documented, the design process by which to realize it is non-trivial \cite{howard2019evolving}. A classical process for designing robotic systems is through the use of models and simulations.  \emph{Model-based} approaches often demonstrate the design of virtual soft robots \emph{in silico} via artificial evolution. By mimicking the processes from which embodied intelligence emerged in nature, these methods have successfully demonstrated complex, meaningful robot behaviors \cite{cheney2013unshackling}. However, evolved robot designs frequently do not perform as expected when transferred into the real world: this is an aspect of the so-called reality gap \cite{mouret201720}. Further still, available fabrication technologies may prevent the realization of virtually evolved robot body-plans altogether. Alternatively, \emph{model-free} approaches restrict design to the physical world, for example via top-down imitation of biological systems such as octopus tentacles or elephant trunks \cite{nakajima2015information} or using real-world evolutionary approaches \cite{vujovic2017evolutionary}. By evaluating robots \emph{in reality} the problems associated with transference between models and the real world can be avoided. However, physical experimentation is generally time-consuming and resource intensive, restricting the effective exploration of large design spaces. 

\begin{figure}[H]
	\centering
	\includegraphics[width=0.8\columnwidth]{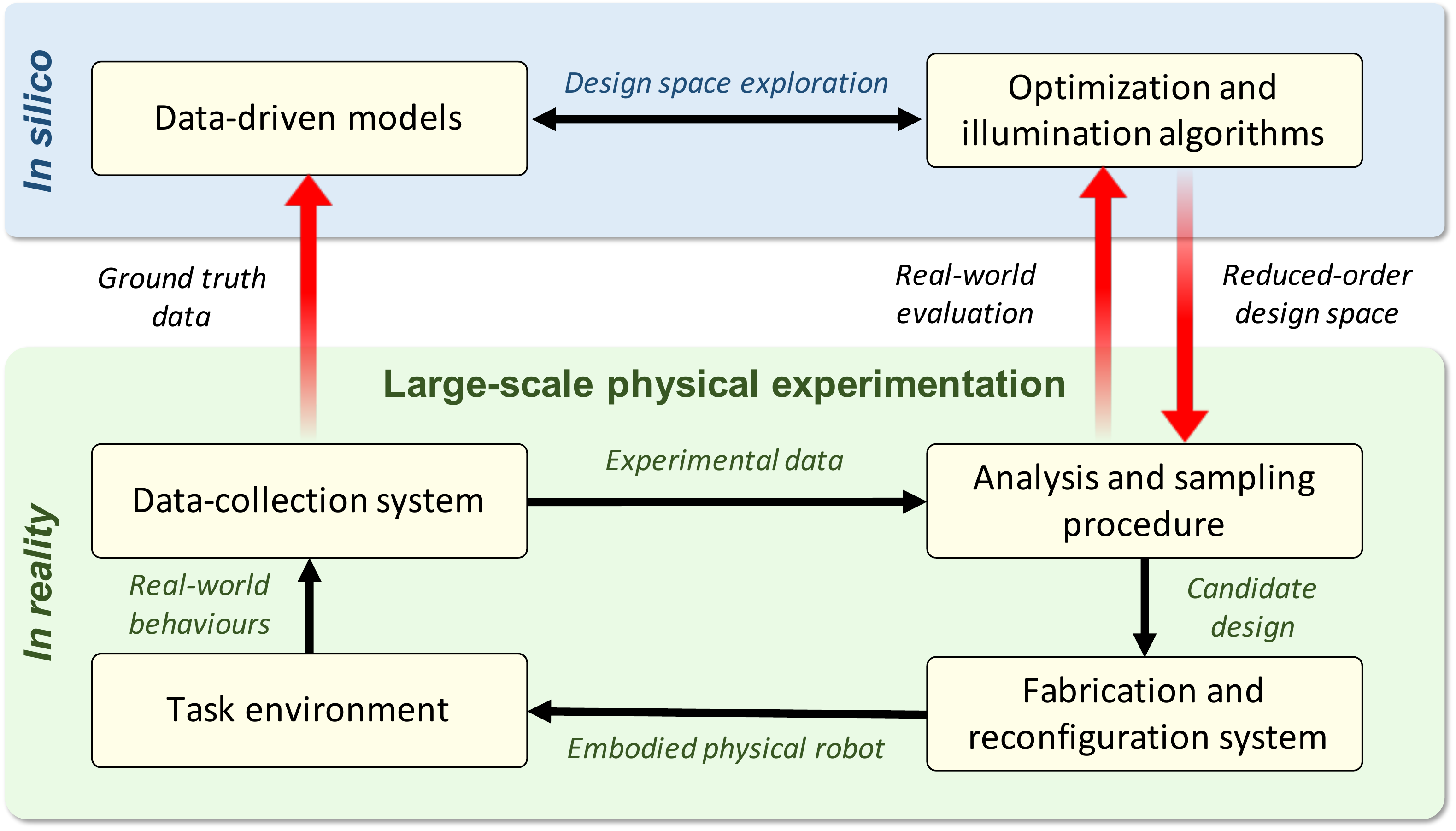}
	\caption{\doublespacing The reality-assisted evolution framework for designing physically embodied soft robots. \emph{In silico}, data-driven models are constructed and adapted based on ground-truth data. Physics engines can be tuned to maximize their predictive accuracy. Alternatively, machine learning can be used to build models from scratch. Auxilary models can also be built to estimate the performance disparity of virtual robots after transference to the real world.  By using data-driven models to simulate huge numbers of possible robot designs, optimization and illumination algorithms can discover a range of diverse and highly performant designs across the parameter space. A reduced-order design space of promising candidates can be established for transference to the real-world. \emph{In reality}, large-scale physical experimentation facilitates the fabrication and analysis of multiple candidate designs. Increasingly automated experimental platforms can fabricated and test large numbers of candidate designs without human input. Alternatively, new modular and adaptive robotic systems facilitate the testing of multiple robot designs on one reconfigurable platform. By placing physical robots in a task environment and observing their behaviors, large volumes of useful experimental data can be gathered. This data can inform the sampling procedure for physical experimentation \emph{in reality}, and can be used to update data-driven models and as an input to optimization algorithms \emph{in silico}. From the bottom up, changes to the physical task environment drive the gradual complexification of data-driven models and robot designs toward more complex embodiments and behaviors. }
	\label{Fig1}
\end{figure}

In this article we summarize recent developments in the design of soft robots from the perspective of a rapidly growing direction we term \emph{reality-assisted} evolution. There has been a growing interest in design methodologies that combine model-based and model-free approaches to improve the overall design of physically embodied robots (see Table \ref{Table1}). Under this framework, as summarized in Figure \ref{Fig1}, simulation-only and simulation-free design methodologies are unified via large-scale physical experimentation and data-driven modelling. \emph{In silico}, data-driven modelling  is used to build and improve representations of the target system based on real-world experimental data. Physics engines can be tuned to maximize their predictive accuracy, and auxiliary models can be built to estimate the transferability of virtual robots to the real world \cite{koos2012transferability}. Alternatively, models can be built from scratch to estimate fitness across the design parameter space \cite{saar2018model}. These data-driven models can be used to simulate the behavior of huge numbers of virtual robots and, in conjunction with optimization and illumination \cite{mouret2015illuminating} algorithms, to discover diverse and highly performant designs that are likely to survive crossing the reality gap. \emph{In reality}, large-scale physical experimentation systems can fabricate and test multiple robot designs. Recent developments in automated \cite{howison2020large} and scalable \cite{kriegman2020scalable} fabrication approaches facilitate an increasingly large number of real-world evaluations. Similarly, modular and adaptive robots allow multiple morphologies and controllers to be tested on a single reconfigurable platform \cite{nygaard2018real,veenstra2018evolution}. By situating physically embodied robots within a task environment, significant amounts of useful experimental data can be gathered. Grounding the design process in physical experimentation ensures the complexity of simulated robot designs does not outpace the model accuracy or available fabrication technologies. Further still, modulating complexity in the task environment \emph{in reality} can drive the heterogeneous improvement of data-driven models \emph{in silico} from simple to more challenging design representations. 

This review is structured as follows. In section \ref{sec2} we outline the open problems in embodied intelligence for soft robots. We highlight three classes of soft robot that typify embodied intelligence in soft robots and drive the need for a reality-assisted framework. In section \ref{sec3} we discuss advances in modelling and representation for soft robotics, starting from conventional approaches and finishing with data-driven modelling and transferability methods. In section \ref{sec4} we discuss advances in large-scale physical experimentation. We cover the range of possible fabrication methods before exploring how these can be scaled up, for example via robotic automation or modular and distributed systems. In section \ref{sec5} we discuss optimization methods for designing soft robots. We exploring design encoding and typical optimization algorithms, then explore how novelty based search methods offer a more robust approach within the reality-assisted framework. Finally, we conclude our review and provide some remarks on outlook.  

\section{Embodied Intelligence for Soft Robots} \label{sec2}

\begin{figure}[H]
	\centering
	\includegraphics[width=0.8\textwidth]{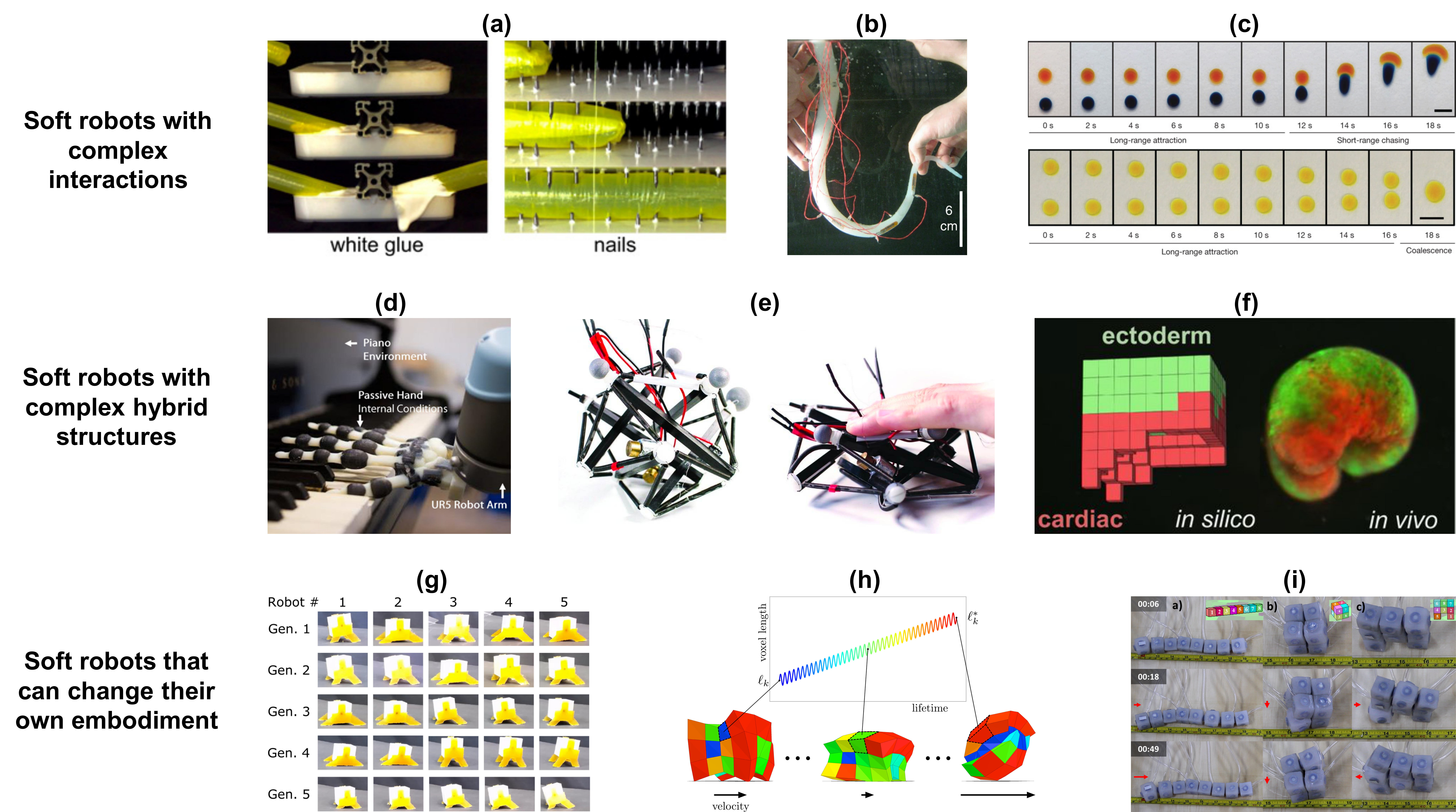}
	\caption{ \doublespacing Three distinct classes of soft robot with strongly embodied functionalities: soft robots with complex interactions, soft robots with complex hybrid structures and soft robots that can change their own embodiment. (a) a soft extendable tube that exhibits complex behaviors via environmental interaction \cite{hawkes2017soft} (b) a simple water-immersed silicon tentacle whose dynamics can be utilized for real-time computation \cite{nakajima2015information} (c) liquid droplets whose properties drive sensing and motility behaviors \cite{cira2015vapour} (d) an anthropomorphic hand with rigid bones but soft ligaments utilizes its hybrid structure to achieve complex dexterity tasks \cite{hughes2018anthropomorphic} (e) a tensegrity robot that combines the robustness of rigid components with the versatility of a deformable body \cite{rieffel2018adaptive} (f) a living tissue robot with hybrid structure for locomotion, where green is passive tissue and red is active contractile cardiac tissue \cite{kriegman2020pipeline} (g) examples of soft robots produced by real-world evo-devo \cite{vujovic2017evolutionary} (h) a simulated soft robot showing the power of morphological development for guiding evolution \cite{kriegman2018morphological} (i) a re-configurable soft modular robotic platform reproducing morphogenetic processes \cite{vergara2017soft}. All figures reproduced with permission from the authors.}
	\label{Figure2}
\end{figure}

Conventional robotic systems can usually be described using a discrete architecture. The exact position of actuation, sensing and other morphological elements is known and can be relied upon not to change during deployment. Interaction with the environment can be actively regulated, for example in pick-and-place robotic arms. The continuum nature of soft robots means they cannot be described using this discrete architecture, but rather as a set of loosely-coupled parallel processes and environmental interactions spanning the microscopic to macroscopic scales. At the microscopic scale, molecular and material interactions induce physical properties such as elasticity or durability, as well as active behaviors such as those seen in electroactive polymers or shape memory alloys \cite{whitesides2018soft}. At the organism level these properties manifest as distinct higher-order behaviors, with environmental interactions inducing highly non-linear dynamics, for example sustained contact forces or turbulent fluid dynamics \cite{katzschmann2016hydraulic}. Further still, at the population level soft robots can be placed in the context of modular and distributed robotics, with interaction between agents driving complexity. In terms of embodied intelligence, therefore, the structure and interdependence of these interactions is key for the emergence of useful, non-trivial behaviors. Given this, we are particularly interested in three classes of soft robot:

\begin{enumerate}
	\item \textbf{Soft robots with complex interactions.} A key area of interest is soft robots with a \emph{simple} embodiment but \emph{complex} environmental interactions. For example, a soft extendable tube that exhibits complex behaviors via contact with the environment \cite{hawkes2017soft} (Figure \ref{Figure2}a), a simple water-immersed silicon tentacle whose dynamics can be utilized for real-time information processing \cite{nakajima2015information}  (Figure \ref{Figure2}b) or liquid droplets whose properties drive sensing and motility functionalities \cite{cira2015vapour} (Figure \ref{Figure2}c). These non-trivial behaviors that emerge via interaction between morphology and the environment represent a fundamental basis on which more complex behaviors can be built built. 
	
	\item \textbf{Soft robots with complex hybrid structures} Another key area of interest is so-called soft-rigid robots. The rapid progression in additive manufacturing has enabled the construction of heterogeneous continuum structures with anisotropic elasticity profiles. For example, an anthropomorphic hand with rigid bones but soft ligaments utilizes its hybrid structure to achieve complex dexterity tasks \cite{hughes2018anthropomorphic} (Figure \ref{Figure2}d). Complex soft structures can also be built using rigid components with soft connections, for example a tensegrity robot that combines the robustness of rigid components with the versatility of a deformable body \cite{rieffel2018adaptive} (Figure \ref{Figure2}e). Robots constructed using living tissues have also shown the potential of hybrid structures for encoding functional behaviors \cite{kriegman2020pipeline} (Figure \ref{Figure2}f).  Similar to biological systems, hybrid robots can diversify their behaviors by complexifying their morphology.
	
	\item \textbf{Soft robots that can change their own embodiment} The final class concerns one of the most fascinating aspects of soft robots, growth and adaptation. One of the key properties separating biological systems from robots is an ability to change ones own embodiment, over the course of a lifetime or in response to the environment. Adaptation has been shown to improve soft robot evolution in real-world \cite{vujovic2017evolutionary} (Figure \ref{Figure2}g) and simulated robots \cite{kriegman2018morphological} (Figure \ref{Figure2}h). Morphogenesis has been demonstrated on modular robotic platforms \cite{vergara2017soft} (Figure \ref{Figure2}i). Understanding how to harness the power of developmental processes is a key step for progressing from simple soft robots to complex, intelligent and perhaps ultimately conscious machines. 
\end{enumerate}

\section{Modelling and Representation} \label{sec3}

Models (mathematical or otherwise logical representations) and simulations (evaluated models) of soft robotic systems facilitate \emph{in silico} experimentation on a scale many orders of magnitude higher than can be achieved \emph{in reality}. This is beneficial for discovering and testing candidate designs before transference to the real world, and for the systematic testing of different design methodologies, e.g. evolutionary algorithms. However, all model representations of the real world have some form of reality gap \cite{mouret201720}. While it may be possible to construct models of physical phenomena, it may not always be possible to evaluate these models in simulation to a sufficient granularity for the purposes of design. This section discusses approaches for modelling soft robots, from conventional methodologies to the latest physics engines and data-driven modelling techniques.

\subsection{Conventional Modelling}
Popular modelling approaches for soft robots are at the level of kinematics or second-order full dynamics \cite{lipson2014challenges}. Such methods involve deriving precise geometric and dynamic expressions, which are in turn used to derive Jacobian kinematics or equations of motion. For example, elastic systems can be modelled as mass-spring lattices or arrays, providing a simple estimation of deformation. Similar approaches include traditional beam-bending and constant-curvature approximations, which often have analytical solutions. Bespoke models can combine these approaches for different modalities of soft body deformation, for example continuum structures with constant-curvature models \cite{della2019control,runge2017framework,webster2010design}. Finite element methods (FEM) are also used for modelling soft systems \cite{zhang2017smoothed}, and have seen particular usage for exploring the design parameter spaces of soft pneumatic actuators \cite{drotman20173d,connolly2015mechanical} and controller verification \cite{zheng2019controllability}. FEM models are generally computationally expensive, requiring a trade-off between simulation time and accuracy \cite{pozzi2018efficient}.  However, work on real-time FEM has shown some promise for the control of soft robots \cite{zhang2016kinematic,duriez2013control}. 

\subsection{Physics Engines and Simulators}

Generalized physics engines that can simulate soft robots as they interact with their environment are growing in popularity. A number of physics engines offer soft body simulators, including Bullet Physics\footnote{\url{https://pybullet.org/}}, CryEngine\footnote{\url{https://www.cryengine.com/}}, MuJoCo\footnote{\url{http://www.mujoco.org/}}, Open Dynamics Engine\footnote{\url{https://www.ode.org/}} and the NASA Tensegrity Robotics Toolkit\footnote{\url{https://github.com/NASA-Tensegrity-Robotics-Toolkit}}. Perhaps the best known physics engine for deformable bodies is Voxelyze\footnote{\url{https://github.com/jonhiller/Voxelyze}}, a voxel-based representation that has been used for a large number of soft robotics studies \cite{cheney2013unshackling,hiller2014dynamic,kriegman2017simulating,kriegman2020scalable} and has become a benchmark on which different design optimization methods are tested (Fig.\ref{Figure2}f,h, for example). A recent gpu-accelerated re-implementation of Voxelyze, voxcraft-sim\footnote{\url{https://voxcraft.github.io/}}, has recently been released.

There have also been recent advances in differentiable soft-simulators, which have the advantage of solvability using gradient-based optimization algorithms. These are particularly efficient for solving optimal control and motion planing problems. However, implementation for soft systems is challenging and currently highly computationally expensive. Examples of soft-body differentiable systems include Chain Queen \cite{hu2019chainqueen} and the differentiable cloth simulator \cite{liang2019differentiable}. The ability to use gradient-based optimization methods could allow for far more computationally efficient optimization and exploration of design. 

Despite developments in soft-body modelling there are still many limitations. Rigid body approximations, FEM and constant curvature models can be effective at predicting the behavior of specific problems, but do not generalize well. Physics engines such as Voxelyze certainly offer a more generalized simulation environment. Additionally, many physics engines are developed by the video game industry, so come with certain technical and financial weight behind them. However, with generality can come a loss of accuracy when compared to problem specific modelling methods. In general, different physics engines offer different trade-offs and capabilities, each suited to different elements of soft robot simulation \cite{silva2012comparative}. 

\subsection{Data-Driven Modelling}

Significant research has been carried out into data-driven modelling techniques that update their representation based on ground-truth data. Data-driven approaches offer the ability to incrementally improve predictions as more data becomes available. Some approaches construct models directly from ground-truth data, for example deriving symbolic equations to describe real-world phenomena \cite{brunton2016discovering,rudy2017data}. Learnt knowledge of fundamental concepts can be gradually combined for understanding more complex systems \cite{schmidt2009distilling}. Other methods detect underlying structure in large data-sets, for example dynamic mode decomposition \cite{schmid2010dynamic}. 

Alternatively, increasingly black-box approaches can be used to estimate the behaviors of soft systems without explicitly discovering the underlying dynamics. Neural networks, for example, have been used for soft robot pose reconstruction \cite{scimeca2019model}, inverse kinematics \cite{giorelli2015neural} or control strategies \cite{choi2018learning}, and are effective without having to explicitly model the mechanisms behind soft deformation. The rapidly growing field of neural ordinary differential equations could see particular usage in data-driven modelling for soft robots \cite{rubanova2019latent}. 

More abstracted methods bypass direct modelling of the underlying dynamics completely, instead learning surrogate models to map between the design space and fitness landscape. Gaussian process (GP) modelling, for example, is a supervised learning technique for solving regression problems \cite{snoek2012practical}. The power of GP models lie in their leveraging of Bayesian inference to estimate black-box function behaviors in relatively few samples. GP models have been used to estimate design fitness in the reality-assisted design of a hopping \cite{saar2018model} robot, mapping morphological and control parameters to locomotion speed, and a tensegrity robot \cite{rieffel2018adaptive}, mapping motor speeds to locomotion speeds.

\subsection{Transferability, Robustness and Sim2Real}
Related to data-driven modelling is the idea of transferability. Simulation-to-real (sim2real) performance can be improved by learning the disparity between robot behaviors observed \emph{in silico} and \emph{in reality} (e.g. their transferability). The \emph{transferability approach} augments existing system models with supervised learning to understand the limits of the model and simulator \cite{koos2012transferability}. This so-called robot-in-the-loop method uses a minimal number of physical experiments to build a surrogate model of the sim2real disparity across the design space. By including this disparity measure within a multi-objective optimization framework, the design fitness landscape can be adjusted to reward designs that are likely to cross the reality gap with a low disparity. Incremental testing in the real world improves the sim2real disparity surrogate model accuracy, driving the design process towards performant \emph{and} transferable solutions. A recent study on a flapping robot, whose dynamics are notoriously challenging to model, used incremental simulated and real-world experimentation to explicitly measure the sim2real disparity across the design space \cite{rosser2019sim2real}.

Robustness filters can be also used to assess how well simulated designs perform under perturbations \cite{jakobi1997evolutionary}, for example in their morphology, controller or environment. If the reality-gap is viewed as a large perturbation to the design space, then designs that can withstand high perturbations in simulation and still exhibit the same behaviors are likely to also perform well in the real-world. These methods have been used for the design of living tissue robots \cite{kriegman2020pipeline}, in which simulated robot designs were passed through a robustness filter that injected noise into their control systems, and a transferability filter that measured their performance once transferred to reality and returned this to the model environment in the form of constraints. Transferability and robustness methods are useful tools because they maximize the utility of existing simulation tools, allowing generalized physics engines to be used in many different design contexts. However, the design process is still fundamentally limited by the accuracy of the model tool. Areas of the design space that cannot be simulated accurately may still provide useful data, but cannot be realized in the form of a physically embodied robot. 

\section{Large-Scale Physical Experimentation}  
\label{sec4}

Those implementing reality-assisted evolution rely on carrying out a sufficient number of physical experiments. In this section we discusses methods and challenges for fabricating and testing physical soft robots on a large scale. First we discuss fabrication, sensing and actuation approaches for soft robots, drawing attention to the difficulty in realizing simulated designs in the physical world. Next we discuss approaches for scaling up experimentation to enable more design evaluations in the real world, including automated fabrication and modular robotic systems. 

\subsection{Materials and Fabrication}

One of the great opportunities offered by soft robotics is the large pool of potential materials and fabrication technologies available to the designer \cite{schmitt2018soft}. Many soft robots are fabricated using casting and molding, typically with silicon elastomers or expanding foam \cite{ilievski2011soft,polygerinos2017soft}. These approaches offer a flexible and potentially scalable approach for fabrication, but can be a highly labor intensive. Additive manufacturing has also recently seen widespread adoption, with 3D printing shown to be especially useful for multi-material soft-rigid robots, e.g. an anthropomorphic hand \cite{hughes2018anthropomorphic} (Fig.\ref{Figure2}d), and for seamless incorporation of sensing and actuation mechanisms \cite{wehner2016integrated}. 3D printing can also be exploited with flexure-based rigid techniques \cite{hughes20173d}, with increasingly high-quality flexible filaments and stereolithography techniques offering significantly more capabilities \cite{gul20183d}. 

There are a number of alternative materials and approaches which potentially offer more scalable approaches.  This includes hot melt adhesives, which have seen widespread usage for soft robots \cite{vujovic2017evolutionary,nurzaman2013active}, and offer the advantage of being cheap, accessible and quick to form. Reconfigurable soft robots are another approach, and can be fabricated using cutting and folding inspired by origami \cite{onal2011towards,rus2018design,zhakypov2018design}. Similarly, laser cutting is a rapid method for fabricating simple robots \cite{howison2020large}, or creating 3D flexure based actuators \cite{lipton2018handedness}. In addition to these well established techniques there are a number of exciting novel fabrication approaches emerging, including knitted or balloon-structure robots \cite{maziz2017knitting,yarbasi2018design}.

\subsection{Sensing and Actuation}
\label{S&A}

Another key challenge in the development of soft robots is sensing and actuation \cite{pfeifer2012challenges}. Actuation methods such as pneumatically-driven systems are popular, however these often require tethers or large external compressors \cite{sun2013characterization}.  Tendon-driven systems are also common and allow for use of motors as an actuation source, providing a higher controllability and power density \cite{camarillo2008mechanics}. While these can be efficient, their size and rigidity makes them hard to integrate into continuum bodies. More unconventional approaches use smart materials, for example dielectric actuators or shape memory alloys \cite{li2019dielectric,motzki2019design}. These methods have the clear advantage that actuation can be seamlessly integrated into a robot morphology and are especially useful from the perspective of embodied intelligence. However, it is currently challenging to generate large forces with these technologies \cite{gu2017survey}. 

Sensing is challenging due to the large deformations and potentially infinite degrees of freedom of deformation that soft-body systems exhibit \cite{shih2020electronic}.  A number of different sensing approaches have been proposed including piezeoresistive materials \cite{firouzeh2015soft}, capacitative sensors \cite{frutiger2015capacitive}, ionic metal soft sensors \cite{chossat2013soft}, and also sensors which use cameras \cite{gilday2020vision} or exploit physical structures \cite{scimeca2019model}. Again, more novel methods have been presented, for example harnessing the sensing properties of liquid droplets \cite{cira2015vapour} (Figure \ref{Figure2}c).

A growing trend in soft robotic sensing and actuation is the use of bio-hybrid materials. The boundaries between biological and artificial systems can be blurred with the development of bio-hybrid robots \cite{romano2019review}.  Some of the capabilities, structures and materials from biological systems can be incorporated to interact with robots \cite{alapan2019microrobotics}.  For example, a bio-hybrid system could incorporate biological muscles, where the muscles are controlled artificially \cite{morimoto2018biohybrid}. This approach could allow the functional gap between current soft-robotics systems and biological systems to be closed.

\subsection{The Fabrication Gap}

The fabrication approaches discussed in \ref{S&A} have a complex set of limitations dictating what types of design can and can't be built to a sufficient level of accuracy. Hence, if the fabrication process is not properly included in the design process we may see a \emph{fabrication gap} \cite{rieffel2005crossing}, i.e. the difference between the prescribed design and physically fabricated end product. The fabrication gap problem can be addressed by considering available fabrication methodologies early in the design process. 

One approach is to use fabrication rules to filter designs at the simulation stage, for example using a build filter to rule out designs that cannot be physically fabricated \cite{kriegman2020pipeline}. Alternatively, designers can use fabrication rules---3D printing commands, for example---as a means to describe designs (e.g. in \cite{rieffel2005crossing}).  \emph{Grammar}-based approaches such as this allows designs to be described via vocabulary of parts and fabrication methods \cite{lau2011converting}. These ideas have recently been advanced with the concept of a \emph{robot compiler}, e.g. a process for converting a robot design into parts, structures and fabrication instructions. Robot compilers have been used to convert simulated designs to origami robots whose fabrication is guaranteed \cite{mehta2014end,mehta2014cogeneration,schulz2017interactive}.

Each fabrication technology introduces its own characteristic to the final designs, which in turn affects their embodiment and thus their behavior. Selecting the correct fabrication approach could vastly increase the performance of soft robots. Indeed, if fabrication is viewed as a \emph{filter} between the virtual and physical environments then this could improve the design process by automatically ruling out designs that could not cross the fabrication gap. The recent work on robot compilers and grammars demonstrates, to some extent, that including fabrication at the representation stage can allow us to harness limited fabrication technologies to our advantage in the design process. 

\subsection{Scaling Up Physical Experimentation}
There is a growing trend for testing large numbers of robot designs \emph{in reality}, and significant research is being carried out to enable more scalable physical experimentation methodologies. Indeed, this is one of the driving factors in presenting the reality-assisted framework as a feasible option for unifying model-based and model-free design methodologies. 

\begin{figure}[H]
	\centering
	\includegraphics[width=0.8\textwidth]{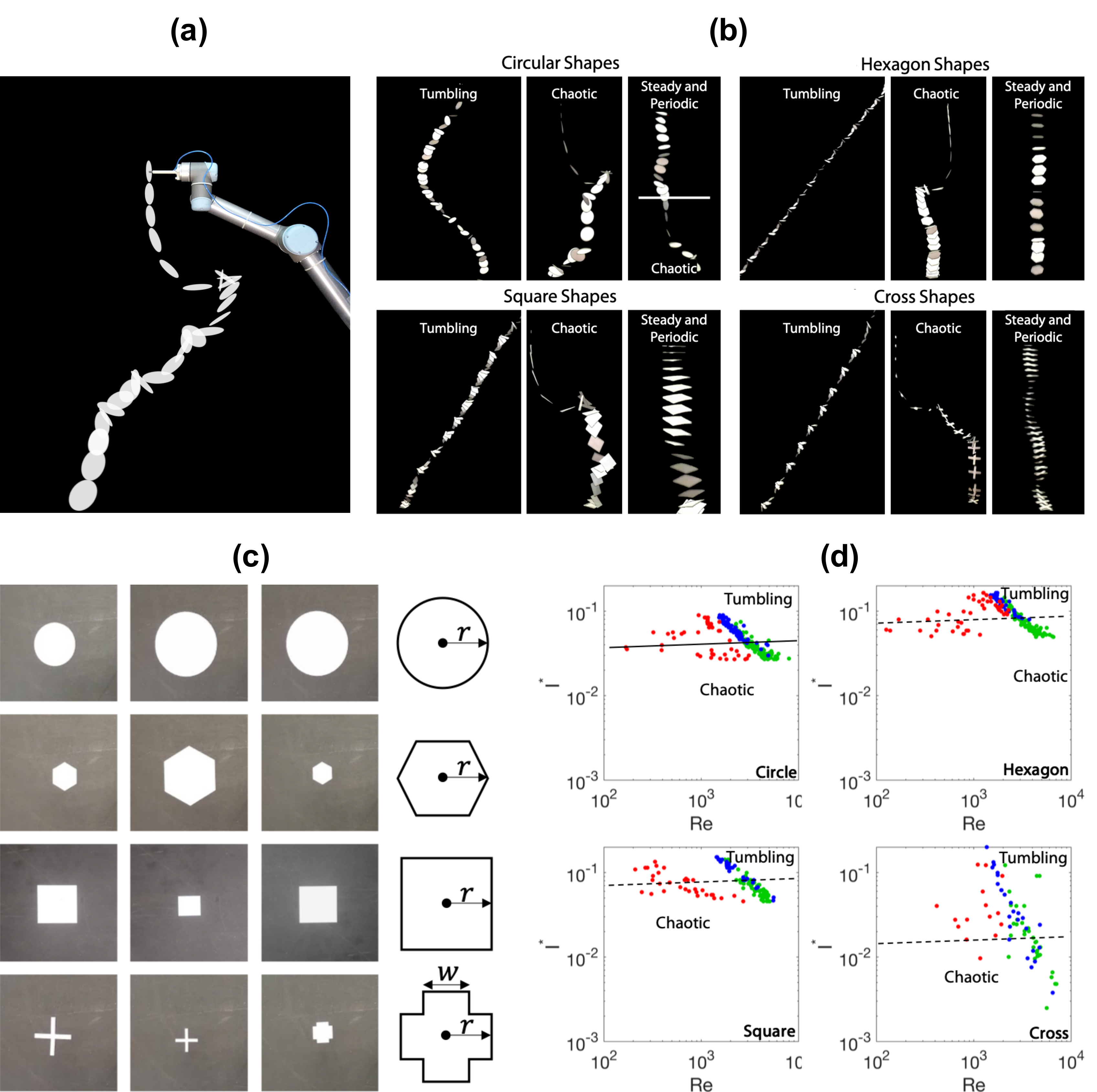}
	\caption{ \doublespacing Large-scale physical experimentation and data-driven modelling with falling paper \cite{howison2020large}. (a) A large-scale physical experimentation system dropping a paper shape. Different paper shapes are automatically fabricated, picked and dropped from a range of initial conditions using a robotic arm. The trajectories of shapes are tracked using computer vision. (b) As the paper shape and initial drop conditions change, different falling behaviors are observed. Falling behaviors are easy to describe visually and simple to investigate experimentally, yet challenging to simulate. (c) Examples of different fabricated morphologies, each with a different design space parameterization (d) Hundreds of trajectories for different shapes are used to create a data-driven model (based on k-means clustering) of behavioral diversity in an appropriate parameter space, in this case Reynolds number (Re) and dimensionless moment of inertia ($I^*$). }
	\label{Fig3}
\end{figure}

Online controller learning is relatively easy to implement with physical experimentation as the morphology remains fixed, and has been applied successfully for locomotion \cite{khazanov2014evolution,nygaard2018real} and a soft fish robot \cite{veenstra2018evolution}. Other studies have co-optimized control and morphology in the real world. One platform uses a ``mother robot" to automatically construct different robot designs and test them for their locomotive behaviors (Figure \ref{Fig3}g) \cite{vujovic2017evolutionary,brodbeck2015morphological,rosendo2017trade}. By automating the fabrication process, large numbers of real-world robots can be evaluated with minimal human intervention. These studies reveal the effectiveness of scaling up physical experimentation, but are still fundamentally limited to orders of magnitude fewer design evaluations than is achievable \emph{in silico}. 

To address this, methods of rapidly fabricating soft robots have been developed, and robotic automation has been used to partially or fully automate the fabrication process.  A number of soft robotic fabrication techniques are particular suited, or can be adapted for rapid fabrication. Additive manufacturing using flexible thermoplastic elastomers \cite{wallin20183d,gul20183d} is one example, as is ``1-D" printing techniques that can fold a single string of recyclable material into a target morphology \cite{cellucci20171d}. In this approach thermoplasters can be extruded using high precision robot arms to create structures, or to connect existing components together.  This approach offers a high level of accuracy and design complexity in a relatively short time. New printing technologies are also developing where complex 3D structures can be fabricated in one step \cite{wehner2016integrated}, improving the speed at which robot designs can be fabricated. Alternatively, low cost soft robot construction kits (voxcraft) have been presented along with appropriate simulation tools to design and test many physical robots \cite{kriegman2020scalable}. Laser cutting is another fabrication approach that can be used to rapidly create bespoke structures with high precision.  When combined with robotic arm automation this can be used to automatically create complex 3D structures. Laser cutting of paper has been shown to facilitate fabrication of morphological structures in under 10 seconds and significantly large-scale physical experiments (Fig.\ref{Fig3}) \cite{howison2020large}. 

\subsection{Modular and Distributed Systems}
Modular robotic systems allow significant physical experimentation without the need to continually fabricate new robots. Instead, modules can be reconfigured to make use of existing fabricated components. Not only does this speed up experimentation, but it allows investigation into bio-inspired topics such as morphogenesis. For example, emergent self-assembly approaches inspired by nature have been replicated in a number of distributed robotic systems \cite{rubenstein2014programmable,werfel2014designing}. Here, global strategies are converted to local assembly rules via a compiler. Modular robot systems such as these are forced to consider fabrication early in the design process, as modules can only be connected in certain configurations. In doing so, users can create an implicitly ``buildable" design space.

Systems where either the assembly or disassembly process is manually performed include for example the programmable-matter systems Robot Pebbles \cite{gilpin2010robot} and Miche \cite{gilpin_miche:_2008}. Even more interestingly, there is a special type of modular robots which can autonomously self-assemble and self-disassemble, i.e. self-reconfigurable modular robots. In theory, these systems could perform experiments autonomously (limited by their operation time), as they would assemble into a design, test it, and re-assemble into the next design by themselves. Within the domain of rigid robots there have been a large number of different such modular robotic technologies. Recent examples of these include M-blocks \cite{romanishin2013m}, SMORES \cite{davey_emulating_2012}, Soldercubes \cite{neubert_soldercubes:_2015},  Roombots \cite{hauser_roombots_2020}. Realizing soft modular or self-assembling robots is less common, largely due to the inherent difficulties in fabricating and controlling continuum bodies.  However, modular soft robots are seeing increasing use \cite{kriegman2020scalable}, for example tensegrity systems \cite{zappetti2017bio}, tendon-driven structures \cite{malley_flippy:_2017}, and soft modular cubes for investigating morphogenetic movements of the embryo (Figure \ref{Figure2}i) \cite{vergara2017soft}. 

A key related field is self-replicating robotics. Introduced half a century ago by John von Neumann, self-replication has long been a dream of roboticists.  Requiring futuristic technologies, it would allow a self-sustaining pipeline of robots to be fabricated autonomously. A number of examples have been developed \cite{suthakorn2003autonomous}, and 3D printers have provided an imperfect replicator \cite{bowyer20143d}.  However, there has not yet been significant progress towards soft self-replicating systems.

\section{The search for novelty} 
\label{sec5}

Key to the success of any design optimization process are the methods used to search for favorable designs and the metrics by which designs are compared. As discussed in \ref{sec4}, a key process in reality-assisted evolution is discovering designs in simulation that are likely to transfer well to reality. In this section we summarize advances in optimization techniques for soft-robot design.  

\subsection{Bio-Inspired Encoding}
Robot designs are typically described via an encoding which systematically maps input parameters to design features. The design space of all possible designs is, therefore, defined by the encoding along with the valid range of input parameter values. There is a range of desirable encoding features. Encoding schemes should output feasible designs for a given task environment, but ideally generalize to many task-environments. They should be receptive to the influence of optimization algorithms searching for favorable designs within the design space, i.e. they should be \emph{evolvable}. Finally, as mentioned previously, having a grounding to the available fabrication technology is advantageous.

Direct encodings map input parameters directly to design features. This can be a powerful approach as it allows a human designer to specify the key features of the output design, hence ensuring the design space is feasible and can be fabricated. For example, robot foot and leg parameters have been optimized with direct encodings \cite{saar2018model}. However, the approach performs poorly in terms of scalability and generality. Since every node in the discretized design space needs a separate, exact description, the total amount of information stored in the direct encoding is as large as the design itself. This can be problematic, especially for soft robots where a fine design discretization can quickly make the optimization process intractable. Furthermore, a direct encoding scheme suitable for a given task may not transfer to another.  

Instead, research has focused on \emph{indirect} or \emph{generative} encoding systems that can describe a complex design space with relatively few parameters \cite{hornby2001advantages,mccormack2004generative}. The idea is inspired by biological systems, where phenotypic expression often occurs in repeating patterns that can be described by a singular part of the encoding, in this case a genome. Lindenmayer systems (L-systems) and their variations \cite{jacob1994genetic,von2007genetic} are generative encodings that incorporate bio-inspired aspects of morphogenesis, the development of morphological characteristics. Their particular properties make them effective for modelling natural plant systems, with directed growth and branching. However, they have not yet found a clear application in the design of soft robots.

Another more promising indirect encoding is \emph{compositional pattern-producing networks} (CPPN's) \cite{clune2011performance}. CPPN's mimic the natural ontogenetic process without the need to directly simulate the chemical mechanisms involved. Instead, describing morphology in terms of a network where each node is a mathematical function (e.g. a Gaussian or cosine function), and the final design is the result of design coordinates queried through this network. By including regularity in the encoding \cite{clune2011performance}, design information can be compressed significantly while maintaining diversity within the design space. CPPN's have been combined with a various optimization algorithms that evolve their underlying network to design soft robots in a simulation environment \cite{cheney2013unshackling}, with some having been transferred to the real world (Fig \ref{Figure2}g). Growing evidence has supported the idea that generative encoding schemes using CPPN's or similar outperform other approaches \cite{richards2014evolving,tarapore2014comparing}. 

\subsection{Optimization algorithms}
Single objective optimization (SOO) strategies aim to maximize performance in a specific behavioral feature or task \cite{trianni2015advantages}, for example locomotion speed. Evolutionary algorithms (EA's) mimic natural selection and offer an effective approach to searching for favorable designs, providing the desired phenotypic behaviors can be encoded within a fitness function. In simulation they have been shown to effectively optimize a range virtual agents \cite{sims1994evolving} including for locomotion \cite{cheney2013unshackling,duarte2017evolution} or navigating confined space  \cite{cheney2015evolving}. EA's have also been demonstrated in the real world, evolving modular robots for locomotion \cite{brodbeck2015morphological,vujovic2017evolutionary} and for controlling a soft fish robot \cite{veenstra2018evolution}.

Bayesian Optimization (BO) \cite{frazier2018tutorial} is an alternative strategy for the global optimization of expensive-to-evaluate black box functions, and has seen significant usage in SOO problems for robotics. BO methods sequentially build and improve a surrogate GP model of the underlying function and use this to efficiently sample the parameter space. Significantly, BO allows users to choose a range of acquisition functions to decide the next sample point, some of which include a penalty for evaluation time. The algorithm, therefore, can search for designs that maximize the predictive accuracy of the GP model and design fitness while also minimizing time spent doing physical experimentation. 

BO has been used for a range of physical robot design optimization tasks including modular robots \cite{rosendo2017trade}, locomotion control strategies \cite{rieffel2018adaptive}, falling paper shape morphologies \cite{howison2020morphologically} and for the co-optimization of morphology and control \cite{saar2018model}, where it was shown to outperform a human designer. In this study, physical robots were iteratively fabricated and tested. After each test, the design fitness was used to update the GP model which, in this case, mapped leg parameters to predicted locomotion distance. BO has also been used in Intelligent Trial and Error (IT\&E), designed to quickly adapt robot controllers to morphological damage by using real-world experimentation to learn optimal control policies \cite{cully2015robots}.

There is a growing trend for implementing multi-objective optimization (MOO) approaches in the evolution of robotic systems. Since intelligent behaviors can rarely be characterized by a single fitness function, MOO offers a framework for characterizing more complex behavioral fitness functions into the design process \cite{trianni2015advantages}. Multi-objective evolutionary algorithms (MOEA's) use multiple objectives to drive the evolutionary process, offering the designer a set of solutions with trade-offs between the various objectives \cite{mouret2008incremental,mouret2012encouraging}. MOEA's are effective for evolving simulated soft robot behaviors, for example in aquatic environments \cite{corucci2018evolving}.  Transferability approaches also use the MOO framework, finding Pareto optimal solutions that balance design fitness and with an estimation of transferability \cite{koos2012transferability}. MOO frameworks are effective for finding a range of possible design candidates, however will not return solutions in areas of the design space that are not Pareto optimal. Hence, many potential candidate designs can be missed, especially in the case of poorly defined objectives. 

\subsection{Quality-Diversity and Illumination Algorithms} 
Designers can incorporate solution novelty within the optimization objectives. Such algorithms aim to actively create divergence in the solution space, rather than converging to one global maxima, for example by formulating the optimization problem to search only for behavioral novelty, i.e. \emph{novelty search} (NS) \cite{lehman2011abandoning,doncieux2019novelty}. The formulation has been shown to outperform traditional objective based searches for solving task-based problems such as maze solving or obstacle avoidance \cite{mouret2012encouraging}, and for the online morphological adaptation of a simulated underwater robot \cite{corucci2015novelty}. In the context of design optimization, this offer an alternative way of evaluating designs based on their behavior, rather than directly on an fitness function. Because many points in the design space may map to the similarly performant behaviors, this approach could significantly improve search efficiency. However, without any direct measure of solution fitness, searches based on novelty can be too divergent to ever present an optimal solution candidate. 

\begin{figure}[H]
	\centering
	\includegraphics[width=0.8\textwidth]{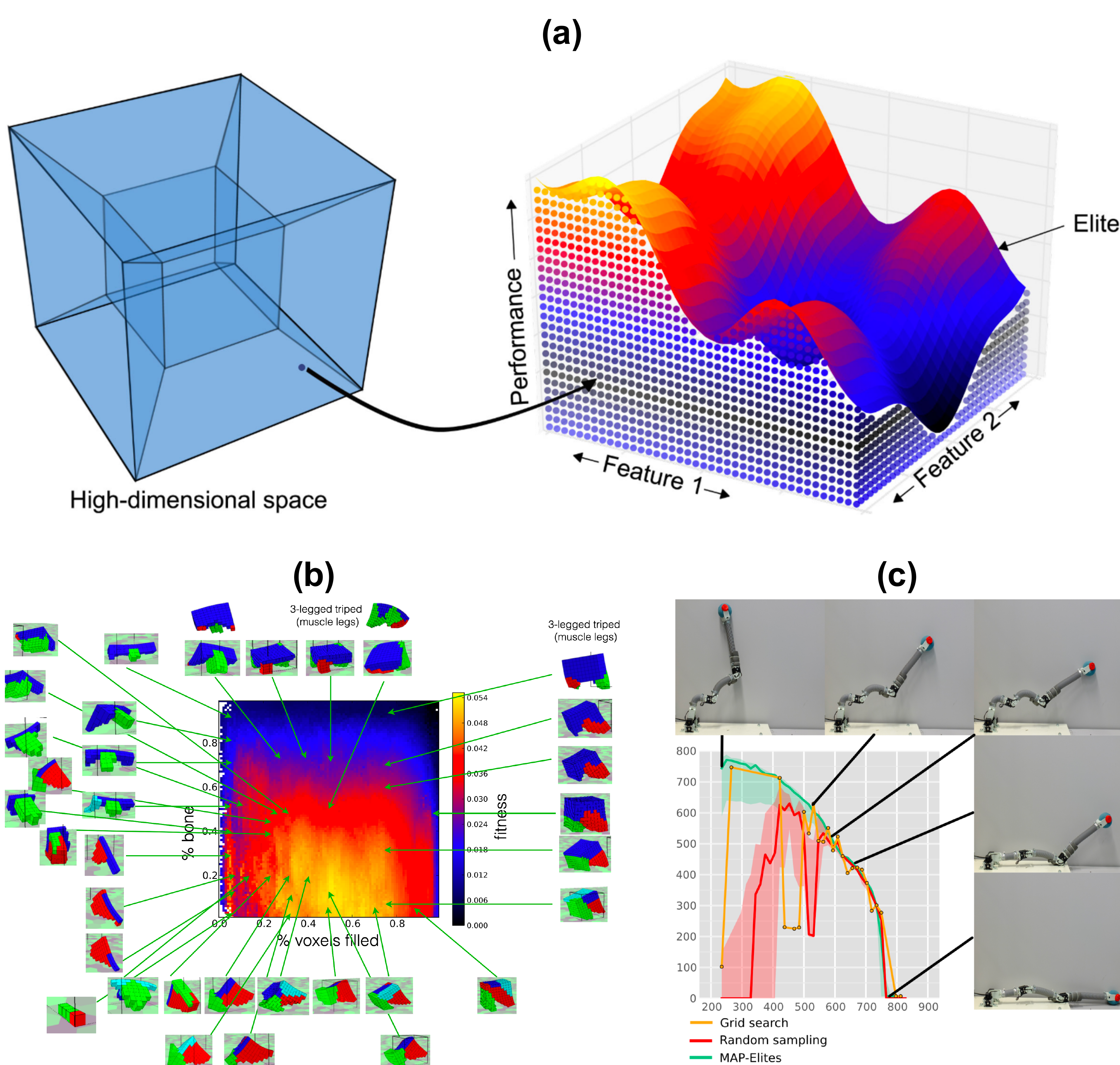}
	\caption{\doublespacing Multi-dimensional Archive of Phenotype Elites (MAP-Elites) algorithm \cite{mouret2015illuminating}. An illumination algorithm that searches for diverse and highly performant solutions across a feature space defined by the user. (a) MAP-Elites searches a high-dimensional parameter space to find highly performant solutions in a user-defined low-dimensional feature space. (b) Examples of different virtual soft robot behaviors found via MAP-Elites. Robots were evaluated for their locomotive fitness across a two-dimensional feature space defined by the morphological parameters (\% bone and \% voxels filled). (c) A soft robotic arm whose controller was discovered using MAP-Elites. The method consistently finds better performing (positive $y$-axis) solutions in the feature space ($x$-axis) than a grid search or random sampling. }
	\label{Fig4}
\end{figure}

To address this, so-called \emph{illumination} or \emph{quality diversity} (QD) algorithms have been presented which hybridize the concept of novelty searching and direct fitness optimization \cite{pugh2016quality}. QD algorithms aim to discover the range of behavioral niches in the system, as well as the associated optimal solution of each niche. One example, Novelty Search with Local Competition (NS+LC) \cite{lehman2011evolving}, lets similarly behaved solutions compete based on a fitness function. When evolving virtual agents for locomotion, NS+LC found more functional morphological diversity than a global fitness function. Similarly, the Multi-Objective Landscape Exploration algorithm (MOLE) \cite{clune2013evolutionary} searches for diverse solutions that are as far away from each other as possible in a user-defined feature space. One of the most popular QD implementations is the Multi-Dimensional Archive of Phenotype Elites (MAP-Elites) \cite{mouret2015illuminating} algorithm (Figure \ref{Fig4}). Here, a set of highly performant solutions are discovered across a discretized feature space defined by the user. The holistic nature of MAP-Elites allows the human designer to define interesting robot behaviors and use these to structure the evolutionary search, and has been reported to outperform traditional EA's in a range of tasks. 

The QD approach may be particularly effective for the design of soft robots because it is more accommodating to unknown system behaviors. In the case of soft robotics, we often do not know how a particular soft system will behave. Optimizing a range of solutions with different behaviors makes it more likely some of these solutions will be effective when transferred into the real world.

\subsection{Developmental Robotics}
Taking inspiration from nature, there is growing evidence that mimicking ontogensis and allowing a developmental phase, where a robot can alter its morphology or controller, can be highly beneficial \cite{rieffel2014growing,doursat2014growing,eiben2013triangle}. Incorporating a developmental process into the design optimization framework adds a number of benefits. First, the design-space complexity can be reduced and offloaded to the developmental interaction between the robot and its environment. Second, performance disparities introduced by the reality gap can be corrected via developmental processes in the real world. Including developmental processes also means a robot can adapt to a changing task environment, or repair itself if damaged. Similarly, controller development allows robots to correct for damage or an unexpected change in task \cite{cully2015robots}.  Third, it could allow the design space of embodied robots to be explored more efficiently and effectively. It has been demonstrated that implementing a developmental framework combined with a novelty search is more effective for evolving simulated soft robots than using a traditional fitness driven approach \cite{joachimczak2015improving}. A key study \cite{kriegman2018morphological} demonstrated how phenotypic plasticity can guide and improve evolution, so-called evo-devo, and increase the overall evolvability across the design space. Large-scale developmental processes have seen limited implementation for real-world embodied soft robots, but the preliminary results indicate the potential \cite{vujovic2017evolutionary}.

\section{Conclusions and Outlook}

This review is written in the context of two key problems in the design of soft robots. First, although model-based methods are effective for searching large design spaces, we are currently unable to sufficiently model many physical processes within soft robots, and are also unable to physically fabricate many virtual soft-robot designs. These reality and fabrication gaps severely limited the scope of designing physically embodied soft robots with a model-based approach. If we subscribe to the view that intelligent behaviors result from complex embodied interaction within the environment, then there is a clear need for significant physical experimentation in lieu of highly accurate modelling and simulation. Second, while model-free physical experimentation does fully capture the embodied behaviors of soft robots, the approach is not scalable to the size achievable with a model-based approach. Designers cannot, for example, explore the evolutionary design of complex soft robots since exploration of these design spaces requires a significant numbers of evaluations. 

The challenges associated with designing soft robots combined with the potential benefits of utilizing soft robots in the context of embodied intelligence are forcing robotiscists to rethink their design methodologies and philosophies. The literature presented here (as highlighted in Table \ref{Table1}) demonstrates that an effective strategy is a reality-assisted evolution framework in which model-based and model-free physical experimentation methods are unified to improve the design of physically embodied soft robots that exhibit useful, meaningful behaviors.\emph{In silico}, data-driven modelling can be used to build and update models of real-world systems. Tuning physics simulations \cite{kriegman2020pipeline}, building models from scratch \cite{saar2018model} and learning auxiliary transferability models \cite{koos2012transferability} are all data-driven methods for reducing the reality-gap disparity. \emph{In reality}, large-scale physical experimentation methods are facilitating significant numbers of real-world evaluations. Automated \cite{howison2020large} and scalable \cite{kriegman2020scalable} methods show promise, as do modular and reconfigurable systems \cite{vergara2017soft,nygaard2018real}. At a high-level, novel optimization techniques can guide design exploration \emph{in silico} and \emph{in reality}. QD and illumination algorithms such as MAP-Elites \cite{mouret2015illuminating} seem particularly suited to modulating the design process by discovering multiple diverse candidate designs to be tested in the real world. Bayesian optimization \cite{rieffel2018adaptive} methods also show promise since they allow designers to include the cost of physical experimentation directly in the objective function. 

Looking forward, we expect to see a continuation of the trend toward a more unified approach to designing soft robots. This will be partially driven by the further development of fabrication, material, sensing and actuation technologies. As these become better understood, so will our ability to include them within models and deploy them via large-scale physical experimentation. In parallel, developments in data-driven modelling and transferability approaches will improve the utility of simulation tools, facilitating better predictions of robot behaviors and, crucially, more informed studies into optimal design methods. In the long term, research on \emph{open ended evolution}, e.g. \cite{lehman2015enhancing,huizinga2018emergence}, that aims to indefinitely evolve systems without converging to a particular solution could be significant. Developments in these search and optimization methods are key for structuring the design process toward discovering diverse behaviors. Finally, moving beyond soft robotics we expect to see a trend toward more complex hybrid systems. Biological systems tend to be neither soft nor rigid, so understanding how the combination of different materials improves embodied functionality is a key milestone toward designing complex, intelligent and perhaps ultimately conscious machines.

\newpage

\newpage

\newpage

\newpage
\begin{table}[H]
	\begin{adjustbox}{center}
		\centering
		\scriptsize
		\begin{tabular}{l|l|l|l|l|l|l}
			
			&  & \textbf{Parameter} & \textbf{Model /} & \textbf{Search} &  \multicolumn{2}{c}{\textbf{Realized Designs:}}   \\
			
			\textbf{Author Reference} & \textbf{Target System} & \textbf{Space} & \textbf{Simulation} & \textbf{Method} & \textbf{Morphologies}  & \textbf{Controllers}  \\
			\hline
			
			& & &  &  & & \\
			\textbf{(a) Model-based} &  &  & & & &   \\
			\hline
			
			\TableCite{hiller2011automatic} & Soft robot & M + C & Voxelyze & EA & 5 & 1 \\
			\hline
			
			\TableCite{caluwaerts2014design} & Tensegrity robot & M + C & NTRT & EA & 1 & 2   \\
			\hline
			
			\TableCite{cellucci20171d} & Recyclable robot & M & ODE & MOEA & 3 & 1 \\
			\hline
			
			\TableCite{peng2018sim} & Robotic arm & C & MuJoCo & DNN & 1 & 1 \\
			\hline

			& & &  &  & &  \\   
			\textbf{(b) Model-free}  & & &  &  & &  \\
			\hline
			\TableCite{khazanov2014evolution}  & Tensegrity robot  & C & --- & EA & 1 & 250  \\
			\hline  
			
			\TableCite{brodbeck2015morphological}   & Modular robot  & M + C & --- & EA & 500  & 500\\
			\hline
			
			\TableCite{mouret2015illuminating} & Soft robotic arm  & C & --- & MAP-Elites & 1 & 640  \\
			\hline  
			
			\TableCite{vujovic2017evolutionary}   & Soft-legged robot  & M + C   & --- & Evo-devo & 75 & 75 \\
			\hline
			
			\TableCite{nygaard2018real} & Quadruped & M + C & --- & MOEA & 192 & 192 \\
			\hline
			
			\TableCite{veenstra2018evolution} & Soft fish robot & C & --- & EA & 1 & 200 \\
			\hline

			& & &  &  & & \\
			
			\textbf{(a) Reality-assisted}  & & &  &  & &  \\
			\hline
			
			\TableCite{koos2012transferability} & Multiple & C & Bullet physics &  MOEA & 2 & 2 \\
			\hline
			
			\TableCite{cully2015robots} & Hexapod & C & ODE & IT\&E & 10 & 10 \\
			\hline
			
			\TableCite{rosendo2017trade} & Modular robot   & M + C & GP & BO & 25 & 25\\
			\hline
			
			\TableCite{saar2018model} & Hopping robot  & M + C & GP & BO & 15 & 40 \\
			\hline
			
			\TableCite{rieffel2018adaptive} & Tensegrity robot & C & GP & BO & 1 & 30 \\
			\hline
			
			\TableCite{rosser2019sim2real} & Flapping robot & M & PYROSIM & MOEA & 16 & 1\\
			\hline
			
			\TableCite{kwiatkowski2019task} & Robotic arm & C & DNN & --- & 2 & 2 \\
			\hline
			
			\TableCite{howison2020large} & Falling paper & M  & KMC & Random & 500 & --- \\
			\hline
			
			\TableCite{howison2020morphologically} & Falling paper & M & GP & BO  &  40 & --- \\
			\hline
			
			\TableCite{kriegman2020pipeline} & Living cell robots  & M & Voxelyze &  EA  & 5  & 1 \\
			\hline
			
			\TableCite{kriegman2020scalable} & Modular soft robot & M & Voxelyze & Exhaustive & 108  & 1\\
			\hline
			
			\multicolumn{7}{l}{\rule{0pt}{4ex}\textbf{Key}} \\
			
			\multicolumn{3}{l}{M: \emph{Morphology}} & \multicolumn{4}{l}{DNN: \emph{Deep Neural Network}}\\
			\multicolumn{3}{l}{C: \emph{Control}} & \multicolumn{4}{l}{EA: \emph{Evolutionary Algorithm}}\\
			\multicolumn{3}{l}{GP: \emph{Gaussian Process}} & \multicolumn{4}{l}{MO: \emph{Multi-Objective}} \\
			\multicolumn{3}{l}{KMC: \emph{K-Means Clustering}} & \multicolumn{4}{l}{MO: \emph{Multi-Objective Evolutionary Algorithm}}\\
			\multicolumn{3}{l}{ODE: \emph{Open Dynamics Engine}} & \multicolumn{4}{l}{BO: \emph{Bayesian Optimization}}\\
			\multicolumn{3}{l}{NTRT: \emph{NASA Tensegrity Robotics Toolkit}} & \multicolumn{4}{l}{IT\&E: \emph{Intelligent Trial and Error}}\\
			\multicolumn{7}{l}{MAP-Elites: \emph{Multi-Dimensional Archive of Phenotype Elites}} \\
			
		\end{tabular}
	\end{adjustbox}
	\caption{\doublespacing Highlighted literature for model-based, model-free and reality-assisted of physically embodied robots, showing the target system, design parameters, modelling and simulation tools, design search method and the number of morphologies and/or controllers that were realized \emph{in reality}. Note that in many cases these publications reported significant numbers of additional experiments, for example manual design comparisons. (a) Model-based approaches. Designs are evaluated using modelling and simulation tools and are not tested in the real-world until the end of the design process. (b) Model-free approaches. Designs are exclusively evaluated in the real-world, without the use simulation or modelling tools. (c) Reality-assisted approach. As categorized in this review, designs are evaluated using a combination of modelling tools and real-world experimentation.} 
	\label{Table1}
\end{table}

\section*{Acknowledgements}
This work was funded by The United Kingdom Engineering and Physical Sciences Research Council (EPSRC) MOTION grant EP/N03211X/2, RoboPatient grant EP/T00519X/1 and grant RG92738 for the University of Cambridge Centre for Doctoral Training. There was additional funding from The Mathworks, Inc.

\end{document}